\documentclass{article}
\usepackage[english]{babel}
\usepackage[letterpaper,top=2cm,bottom=2cm,left=3cm,right=3cm,marginparwidth=1.75cm]{geometry}
\usepackage[round]{natbib}
\usepackage{amsmath}
\usepackage{graphicx}
\usepackage[colorlinks=true, allcolors=blue]{hyperref}
\usepackage{microtype}

\title{The language of sounds unheard: Exploring musical timbre semantics of large language models}

\author{Kai Siedenburg$^{*a}$, Charalampos Saitis$^{*b}$\\
$^a$Dept. of Medical Physics and Acoustics, University of Oldenburg\\
$^b$Centre for Digital Music, Queen Mary University of London\\
$^*$equal contribution}

\providecommand{\keywords}[1]
{
  \small	
  \textbf{\textit{Keywords---}} #1
}

\begin{document}

\maketitle

\begin{abstract}
Semantic dimensions of sound have been playing a central role in understanding the nature of auditory sensory experience as well as the broader relation between perception, language, and meaning. Accordingly, and given the recent proliferation of large language models (LLMs), here we asked whether such models exhibit an organisation of perceptual semantics similar to those observed in humans. Specifically, we prompted ChatGPT, a chatbot based on a state-of-the-art LLM, to rate musical instrument sounds on a set of 20 semantic scales. We elicited multiple responses in separate chats, analogous to having multiple human raters. ChatGPT generated semantic profiles that only partially correlated with human ratings, yet showed robust agreement along well-known psychophysical dimensions of musical sounds such as brightness (\emph{bright--dark}) and pitch height (\emph{deep--high}). Exploratory factor analysis suggested the same dimensionality but different spatial configuration of a latent factor space between the chatbot and human ratings. Unexpectedly, the chatbot showed degrees of internal variability that were comparable in magnitude to that of human ratings. Our work highlights the potential of LLMs to capture salient dimensions of human sensory experience.
\end{abstract} \hspace{10pt}


\section{Introduction}
It has recently become apparent that language processing based on large language models (LLMs) is not only here to stay, but also about to change research and society. 
This insight dawned most strikingly with the  surfacing of chatbot \emph{ChatGPT}\footnote{\url{https://openai.com/blog/CGPT/}} (CGPT, \emph{Chat Generative Pre-trained Transformer}), based on the LLM \emph{GPT3} from by the company OpenAI\footnote{\url{https://openai.com}}. 
CGPT mimics a human conversationalist and seemingly covers any type of topic. 
A major line of critique of CGPT yet concerns its relation to facts for it often provides answers that sound plausible but are incorrect. 
For perception research, an important question concerns the extent to which LLMs faithfully recover dimensions of human experience. 
Specifically, the extent to which the perceptual semantics of LLMs map onto human perceptual semantics remains unclear. 
\cite{marjieh2023language} observed that GPT3 structurally aligns with human auditory experience, as probed with dissimilarity judgments of consonants, timbre, pitch, and loudness. 
Here, we wish to explore the extent to which the chatbot CGPT aligns with human judgments of fine grained  semantic dimensions of sound.  

Although auditory sensory experience is rich, humans lack a rich generic vocabulary for sound. In fact, most Western languages have relatively few generic adjectives to describe auditory experience (e.g., soft--loud, shrill; hereafter we will use `--' to indicate semantically opposite and `/' to indicate semantically similar adjectives). Most other adjectives useful for describing sound, for example, high--low, dark--bright, or rough--smooth, are based on crossmodal associations with other sensory or experiential modalities \citep{wallmark2018corpus, saitis2020timbre}. 
In fact, research has demonstrated pronounced crossmodal associations between sound and other modalities such as touch that extend across cultures \citep{cwiek2022bouba, winter2022trilled}. 
The study of semantic dimensions of sound has therefore important implications for understanding the relation between perception, language, and meaning. 

When it comes to assessing sounds along semantic scales, studies have repeatedly found low dimensional factor spaces with around three dimensions \citep{zacharakis2014interlanguage, saitis2019semantics}. Factors are latent variables resulting from dimensionality reduction techniques such as exploratory factor analysis and principal components analysis \citep{osgood1952nature}. 
This finding suggests either a limited dimensionality of sound semantics or limited agreement among participants. 
Little is known about the internal consistency of chatbots such as CGPT. Users quickly grasp that identical prompts can give rise to different responses, but the degree of this variability remains to be quantified. 

In this work, we probed CGPT using semantic scales from a recently published study of sound semantics using musical instrument tones. 
Specifically, we had the chatbot provide ratings of sounds along semantic scales, allowing us to directly compare data from human participants with the chatbot. 
Collecting multiple complete datasets from the chatbot allowed us to compute the consistency of responses. As described in the following, it turned out that the chatbot only agreed with human raters for a selected number of scales. Its internal consistency was remarkably low and in fact comparable to the inter-subject consistency of humans.

\section{Methods}

\subsection{Human Dataset}

We considered a recently published dataset by \cite{reymore2023timbre}: 540 online participants rated notes from eight Western orchestral instruments (see Fig. 1) across three registers (low, medium, and high) on 20 semantic scales (5-point). 
Each scale is described by up to three words: 1) \emph{deep, thick, heavy}; 2) \emph{smooth, singing, sweet}; 3) \emph{projecting, commanding, powerful}; 4) \emph{nasal, buzzy, pinched}; 5) \emph{shrill, harsh, noisy}; 6) \emph{percussive (sharp beginning)}; 7) \emph{pure, clear, clean}; 8) \emph{brassy, metallic}; 9) \emph{raspy, grainy, gravelly}; 10) \emph{ringing, long decay}; 11) \emph{sparkling, brilliant, bright}; 12) \emph{airy, breathy}; 13) \emph{resonant, vibrant}; 14) \emph{hollow}; 15) \emph{woody}; 16) \emph{muted, veiled}; 17) \emph{sustained, even}; 18) \emph{open}; 19) \emph{focused, compact}; 20) \emph{watery, fluid}.

These scales were derived from interviews and rating tasks for imagined typical sounds of instruments \citep{reymore2020using}. They are thus well suited to the task: CGPT cannot listen to a musical tone but might have a general understanding of how it sounds. Indeed, after providing its ratings, CGPT added that these “are based on general observations and typical characteristics of a [instrument]’s sound in different registers.” 

\subsection{CGPT Dataset}

We designed the following prompt template for CGPT:
\emph{Please rate how well the descriptions provided in the following (separated by semi-colons) apply to the sound of a [instrument] for a low-register, mid-register, and high-register note on a scale from 1 to 5 where 1 is "does not describe at all", 3 is "describes moderately well" and 5 is "describes extremely well": 1) deep, thick, heavy; 2) smooth, singing, sweet; ... 20) watery, fluid. Provide your response in matrix form, with columns corresponding to registers (low, mid, high) and rows corresponding to attributes 1-20 (in matched order).} The last bit was necessary to ensure that the model both provided numerical values and in a format suitable for data collection. 

We prompted CGPT in early and middle February 2023 using its online public interface\footnote{\url{https://chat.openai.com/}}.  
For each triplet of stimuli (one instrument $\times$ three registers) 
we collected 50 ratings from CGPT in an equal number of separate conversations with the chatbot, considering each conversation as a unique ``rater.'' 
This ruled out CGPT remembering past ratings---the model is not able to access past conversations to inform its responses to the current conversation---while at the same time allowing us to assess the internal consistency of its ratings and compare it with human ratings. 
Data from four CGPT “raters” were discarded due to many NAs.

\subsection{Analysis}

To evaluate how CGPT assesses sound semantics, we first computed Pearson correlation coefficients between the model's ratings and those made by humans. Subsequently, we quantified the reliability of the model by computing Pearson correlations between all pairs of responses within each group.
In addition, we used exploratory factor analysis to compare latent semantic representations between humans and CGPT. 

\section{Results}

To assess the role of musical training in our sample of human participants, we correlated the average rating profile from musician and nonmusician participants.
In the original study of \cite{reymore2023timbre} musical training was self-reported using the single-question item from the Ollen Musical Sophistication Index \citep{ollen2006criterion}: 81\% of the participants self-identified as nonmusicians (``non-musician'' and ``music-loving non-musician'') and 19\% as musicians (``amateur,'' ``serious amateur,`` ``semiprofessional,'' and ``professional'').
Average ratings from these two groups of participants turned out to be highly correlated (Pearson's $r = .97, CI: [.96, .97], p < .001$). For that reason, all 540 human participants were considered jointly in the following.

\begin{figure}
    \centering
    \includegraphics[width = \textwidth]{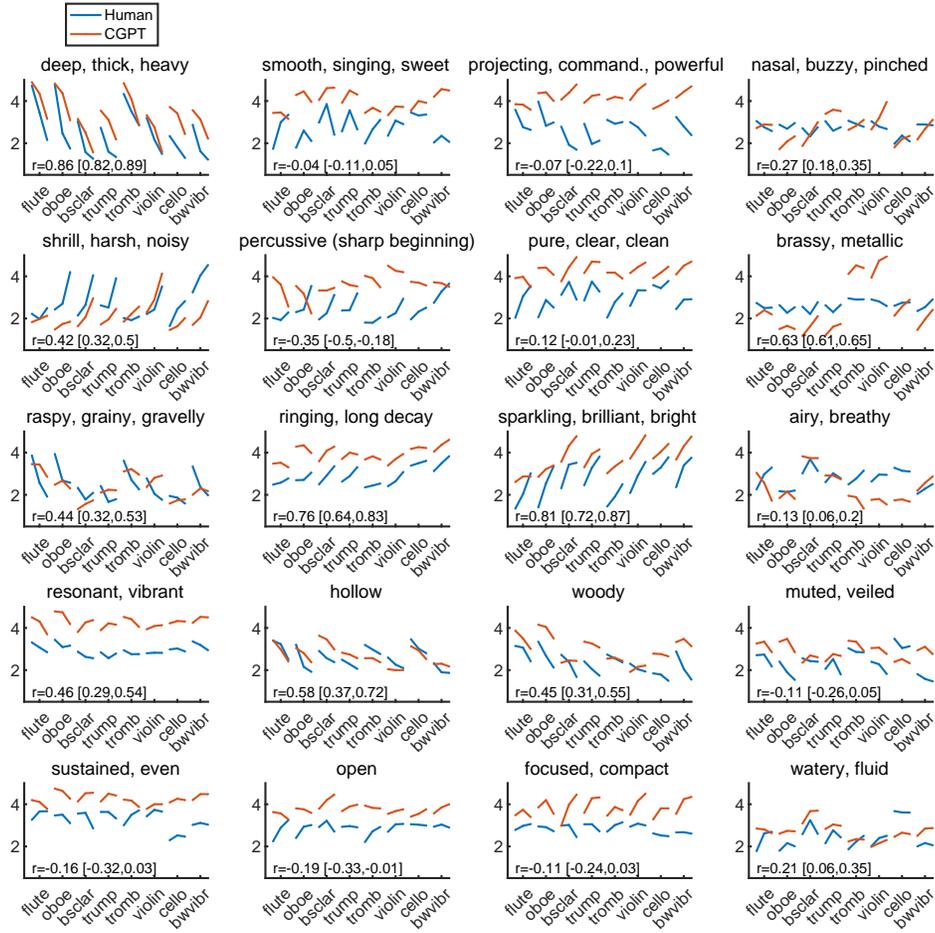}
    \caption{Profiles of average ratings across all 24 stimuli for human and machine ratings.}
    \label{fig:profile}
\end{figure}

An important first comparison between human and machine ratings concerns the average rating profiles. 
Figure~\ref{fig:profile} shows the average rating profiles, with varying degrees of agreement between human and CGPT ratings. 
For instance, the scale \emph{deep, thick, heavy} yielded a strong correlation  of $r=.86 \ [.82, .88]$ between average human and machine ratings (square brackets correspond to 95\% confidence intervals obtained by bootstrapping the CGPT data). 
With somewhat less agreement, CGPT ratings for the scale \emph{sparkling, brilliant, bright} with $r = .81 \ [.73, .87]$ showed similar patterns across registers  but a marked offset towards higher rating points compared to human ratings. 
Overall, twelve scales yielded positive correlations between human and CGPT ratings with confidence intervals non-overlapping with zero (deep, nasal, shrill, brassy, raspy, ringing, sparkling, airy, resonant, hollow, woody, and watery).

\begin{figure}
    \centering
    \includegraphics[width = \textwidth]{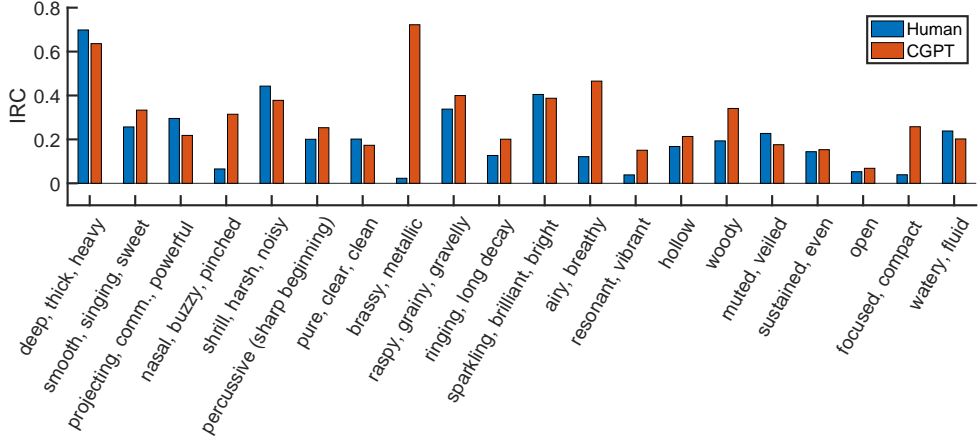}
    \caption{Median inter-rater correlations (IRC) for human and CGPT responses for all 20 semantic scales.}
    \label{fig:IRC}
\end{figure}

Next, we considered the internal consistency of human and machine ratings. 
For every scale, we computed the Pearson correlation between the individual rating profiles of all pairs of participants. 
The resulting distributions of inter-rater correlations (IRC) correspond to the consistency of responses within the two participant groups. 
Figure~\ref{fig:IRC} shows the  distributions' median values. 
Note that we could also have shown the IRC for pairs of human and CGPT responses, but these information are  redundant (and strongly correlated, r(19) = .79) with the correlations between average rating profiles presented above.

The highest IRC for humans was observed for the \emph{deep, thick, heavy} scale, with similar IRC for CGPT responses. 
An interesting case is the \emph{brassy, metallic} scale, with an IRC of 0.72 for CGPT responses, but an IRC of close to zero for humans. 
More generally, calculating the median IRC across scales suggested highest consistency among CGPT responses (0.26) and somewhat lower values for human responses (0.20).  Wilcoxon signed rank tests suggested that the medians of the distribution of human-human IRCs (allocated across scales) was only marginally smaller compared to the CGPT-CGPT IRCs ($z = -1.9, p = .057$).
That is, ratings from CGPT were similarly consistent  compared to human ratings. 

To estimate a baseline below which median IRC values could be considered indistinguishable from chance, we conducted a bootstrapping analysis. 
Median IRC values were computed as before but with randomly shuffled stimulus indices for every participant, using 400 bootstrap iterations. 
We selected the 99th percentile of the resulting bootstrap distribution of median IRC values as a baseline, amounting to baseline values of 0.0001 and 0.022 for human and CGPT responses, respectively. 
This implied that indeed all IRC values from all human  and CGPT responses were above baseline.
Accordingly, even though many of the depicted IRC values are of very small magnitude, they unlikely arose by chance. 

We further considered the variance of the average rating profiles and its relation to the IRC. 
Comparing the average rating profiles and the median IRCs across all 20 scales, we in fact  observed very strong correlations between the standard deviations of average rating profiles and the median IRCs within both types of groups (human: $r = .99, p < .001$; CGPT: $r = .98, p<.001$). This implies that in this data set, average rating profiles with little variance were due to inconsistency among participants and not due ratings with little rating variance across stimuli. 


\begin{figure}
    \centering
    \includegraphics[width = \textwidth]{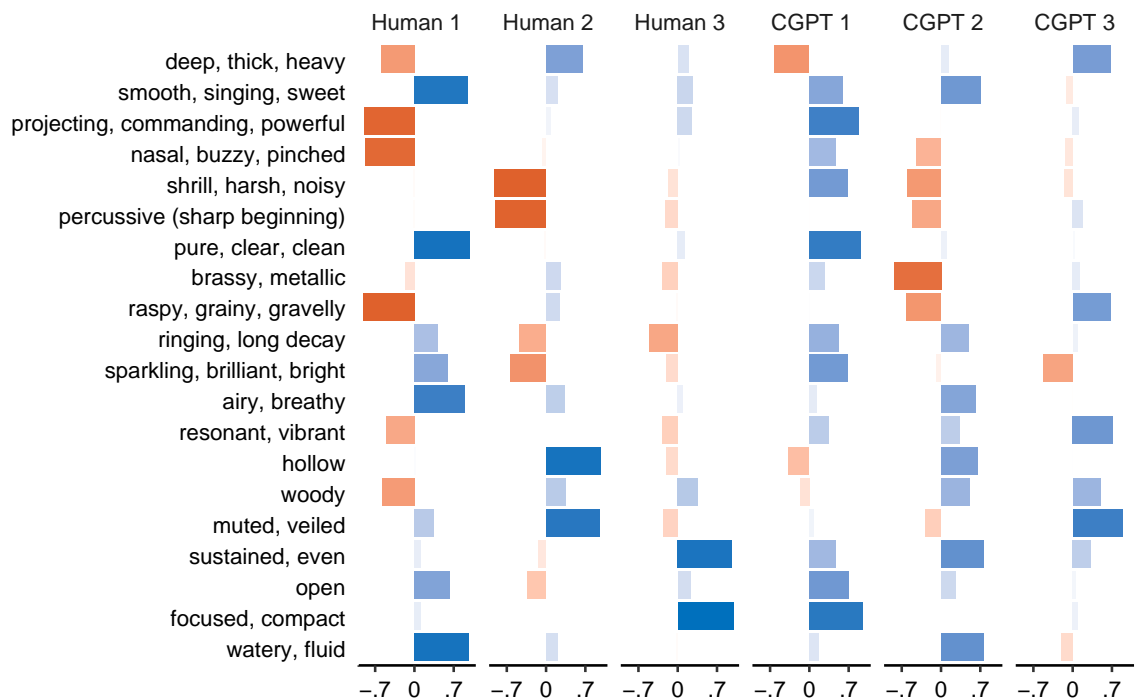}
    \caption{Human and CGPT factor loadings of semantic scales after oblimin rotation.}
    \label{fig:loadings}
\end{figure}

Finally, we examined the structure of human and CGPT representations of sound semantics using exploratory factor analysis based on all 20 scales.
To select an appropriate number of factors for each dataset, we used a bootstrap method known as Horn’s parallel analysis \citep{horn1965rationale}.
Mardia's test for multivariate normality suggested non-normal distributions in both datasets, therefore factor analysis was performed using principal axis factoring \citep{fabrigar1999evaluating} with non-orthogonal oblimin rotation. 
Parallel analysis supported a three-factor solution for both human and machine ratings.
The factors cumulatively accounted for 82\% and 70\% of data variance in the human and CGPT ratings, respectively.
Individual factor variance is not available for the rotated solution due to the non-orthogonality of the factors.

The loadings of factors onto semantic scales are shown in Fig.~\ref{fig:loadings}. 
Correlations between individual human and machine factors were moderate to weak, absolute Pearson $r$ values being between $.003$ (Human 3-CGPT 3) and $.38$ (Human 3-CGPT 2) with confidence intervals overlapping with zero. The overall correlation between the two semantic spaces was indistinguishable from chance, $r = -.07 \ [-.17, .25]$. 
That is, even though robust correlations between human and machine ratings were observed for selected scales, there was no correlation between the overall spatial configuration of all 20 scales as derived by factor analysis.

\section{Discussion}
Here, we took first steps into exploring commonalities and differences in the way in which humans and the chatbot CGPT assess sounds semantics. Specifically, we asked CGPT to perform a recently published rating task of musical instrument sounds along 20 verbally anchored scales \citep{reymore2023timbre}.   
Average human and CGPT profiles correlated significantly for twelve out of twenty scales. 
Specifically, the scales \emph{deep, thick, heavy} and \emph{sparkling, brilliant, bright} yielded strong correlations ($r>.8$). 
It is worth mentioning that these two scales are related to two of the most salient perceptual dimensions of musical sounds, namely pitch height (\emph{deep--high}) and brightness (\emph{bright--dark}). 
These dimensions are not completely independent for natural sounds \citep{siedenburg2021spectral, russo2005interval}, but appear to yield robust agreement between human and CGPT ratings. This is consistent with the findings regarding pitch height by \cite{marjieh2023language}. 
Other less established and potentially less salient semantic scales (e.g., \emph{projecting, ...}, or \emph{muted}) accordingly did not yield robust correlations between human and machine responses. 
Overall, this result suggests a partial agreement between the semantic assessment of sounds by human and CGPT. 
This was further corroborated by an exploratory factor analysis of the raw human and CGPT semantic ratings data. We found that between the two groups the latent semantic representation captured by the 20 scales shared the same dimensionality but dimensional similarity was as good as chance.

A specific feature of our current approach was to gather multiple responses from CGPT, allowing us to quantify and compare the consistency between responses. 
We quantified the amount of variability  between human raters as well as CGPT responses to the same prompt. 
This approach showed highest consistency between responses generated by CGPT with higher levels of IRC of the ratings of the chatbot (across different conversations) compared to human ratings (across different participants). 
Certainly, the reliability of CGPT may be affected by several boundary conditions. For instance, it is specifically designed for chatbot applications, while GPT3 and its successor GPT4 are more general-purpose models and can be used for a wider range of tasks \citep{marjieh2023language,hansen2022semantic}. 
Further, the online public CGPT interface, which we used to prompt the chatbot, is more diverse and unpredictable in its output than GPT3 due to a default sampling ``temperature'' of 1 versus 0.7. This parameter controls how deterministic or ``creative'' the generated responses are with values between 0 and 1, respectively. (As of early March 2023 it is now possible to specify the CGPT temperature via its API.) It remains an open question to what extent temperature affects the consistency between multiple responses from GPT models when replicating human perceptual tasks \citep[cf.][]{miotto2022gpt}.  
More generally, it seems to be important for  our understanding of CGPT that its ratings show a comparable variance across conversations as responses across human raters. 

The pace of development of large language models such as CGPT is extraordinarily high and by the time of writing this manuscript, the model GPT4 has been released. 
Therefore, the present work can only be considered as a snapshot into how one specific version of CGPT (GPT3, probed in February 2023) construes sound semantics. Yet, researchers might start to track the evolution of machine behavior over time (or development cycles), so that a more complete picture concerning the relation of human and machine semantics can be drawn. 
Analysing the structure and behavior of deep-learning based algorithms optimized to replicate human perceptual tasks has already become an important approach in perception science \citep{eickenberg2017seeing, kell2018task, giordano2023intermediate}. 

An important difference between the present study and that of \citet{marjieh2023language} concerns the relation between the prompted task and what the model might already know about it. 
CGPT and GPT3 are both part of a series of models trained on a blend of text and code from before Q4 2021. The human data considered in this study appeared in early 2023, therefore they do not form part of CGPT's ``knowledge.'' Whereas the human loudness and timbre dissimilarity datasets considered by \citet{marjieh2023language} were published in 1978 and 2018, respectively. 
Is it possible that in their case GPT3 was aware of these studies? And could that have influenced the high correlation they observed for loudness and the moderate correlation for timbre? 

Yet another question that could be explored concerns the design of the prompt. Here, we asked the chatbot to provide several ratings as part of one response in order to keep efficiency of data collection on a reasonable level. From a formal viewpoint, however, one could argue that the collection of one data point per prompt may have better fitted with the human experiment. 
An insurmountable disparity between the way in which data were collected from humans and the chatbot lies in the fact that humans listened to sounds whereas CGPT was simply informed about the instrument name and pitch register. It should be noted that the semantic scales used here were designed on the basis of an auditory imagery task instead of an actual listening task \citep{reymore2020using}. 
Also, even human judgements of sound similarity have been shown to be highly conflated with knowledge about instrument categories and the means of sound production \citep{siedenburg2016acoustic, saitis2020brightness} and dissimilarity relations between imagined and heard sounds strongly correlate \citep{halpern2004behavioral}.  

In summary, the present study addresses the extent to which human sensory judgments can be recovered by machine semantics. 
It turns out that CGPT agrees with humans with its ratings  of a majority of scales including the perceptually most salient ones. 
In his taxonomy of listening, \cite{schaeffer2017treatise} suggested  four distinct modes of listening: \emph{to listen} (écouter), \emph{to perceive aurally} (ouïr), \emph{to hear} (entendre), \emph{to understand} (comprendre). 
In the era of artificial intelligence, one might be tempted to suggest yet another mode that captures what CGPT is doing rather successfully already: \emph{to pretend to understand}. 
The fundamental gap between a listening participant and a non-listening machine remains, but this gap might be closed by future work on machine listening.

\section*{Acknolowledgements}

KS was supported by a Freigeist Fellowship of the Volkswagen Foundation. 
The authors thank Lindsey Reymore for making the data of her original study openly available.  The authors also thank Elif Özgür for helping collect some of the CGPT data. 


\end{document}